\pdfoutput=1

\documentclass[11pt]{article}

\usepackage{ACL2023}

\usepackage{times}
\usepackage{latexsym}

\usepackage[T1]{fontenc}

\usepackage[utf8]{inputenc}

\usepackage{microtype}

\usepackage{inconsolata}

\usepackage{graphicx,xcolor}  %
\usepackage[whole]{bxcjkjatype} %
\usepackage{multirow}
\usepackage{booktabs} %
\usepackage{siunitx}
\usepackage{enumitem}

\usepackage{float}
\usepackage{comment}

\usepackage{soul} %

\usepackage{tikz}
\usepackage{tcolorbox}
\tcbuselibrary{skins}
\tcbuselibrary{breakable}

\newtcolorbox{simplebox}{
  enhanced,
  drop fuzzy shadow,
  colback=white,
  boxrule=0.5pt
}

\newtcolorbox{expbox}{
  enhanced,
  colback=white,
  boxrule=0.5pt,
  breakable
}

\newtcolorbox{alertbox}{
  enhanced,
  drop lifted shadow,
  sharp corners,
  colback=white,
  colbacktitle=white,
  coltitle=black,
  boxrule=1.0pt,
  breakable
}

\newtcolorbox{alertboxa}[1]{
  title={#1},
  enhanced,
  drop lifted shadow,
  sharp corners,
  colback=white,
  colbacktitle=black!10!white,
  coltitle=black,
  boxrule=1.0pt,
  breakable
}

\newtcolorbox{quoteboxa}[1]{
  title=#1,
  enhanced,
  colback=white,
  colbacktitle=white,
  coltitle=black,
  boxrule=0.8pt,
  attach boxed title to top left={xshift=3mm, yshift*=-\tcboxedtitleheight/2},
  box align=top
}

\newcommand{\syllogism}[3]{
\begin{tabular}{l}
\textbf{P1}: #1\\
\textbf{P2}: #2\\ \hline
\textbf{C}: #3
\end{tabular}
}

\newcommand{\rev}[1]{\textcolor{black}{#1}}

\title{
Exploring Reasoning Biases in Large Language Models Through Syllogism: Insights from the NeuBAROCO Dataset
}

\author{Kentaro Ozeki\textsuperscript{1,2}, Risako Ando\textsuperscript{1}, Takanobu Morishita\textsuperscript{1}, Hirohiko Abe\textsuperscript{1},\\ \textbf{Koji Mineshima\textsuperscript{1}, Mitsuhiro Okada\textsuperscript{1}}\\
  \textsuperscript{1}Keio University, Tokyo, Japan\\ %
  \textsuperscript{2}University of Tokyo, Tokyo, Japan\\ %
  \texttt{kentaro.ozeki@gmail.com}\quad
  \texttt{\{risakochaan,morishita,hirohiko-abe\}@keio.jp}\\
  \texttt{\{minesima,okada\}@abelard.flet.keio.ac.jp}}

\begin{document}
\maketitle
\begin{abstract}
This paper explores the question of how accurately current large language models can perform logical reasoning in natural language, \rev{with an emphasis on whether these models exhibit reasoning biases similar to humans.}
Specifically, our study focuses on syllogistic reasoning, a form of deductive reasoning extensively studied in cognitive science as a natural form of human reasoning.
We present a syllogism dataset called NeuBAROCO, which consists of syllogistic reasoning problems in English and Japanese.
This dataset was originally designed for psychological experiments to assess human reasoning capabilities using various forms of syllogisms.
Our experiments with leading large language models indicate that these models exhibit reasoning biases similar to humans, along with other error tendencies.
Notably, there is significant room for improvement in reasoning problems where the relationship between premises and hypotheses is neither entailment nor contradiction.
We also present experimental results and in-depth analysis using a new Chain-of-Thought prompting method, which asks LLMs to translate syllogisms into abstract logical expressions and then explain their reasoning process.
Our analysis using this method suggests that the primary limitations of LLMs lie in the reasoning process itself rather than the interpretation of syllogisms.

\end{abstract}

\section{Introduction}
\label{sec:intro}

The ability to perform accurate deductive reasoning in natural language, once pursued by classical symbolic AI, has recently become a vital issue \rev{in the development and evaluation of Large Language Models} (LLMs)~\cite{qiao2022reasoning,huang2022towards}.
Regarding humans, empirical research in cognitive science has demonstrated that humans often exhibit various errors and tendencies in reasoning, known as \emph{reasoning biases}~\cite{evans1989bias,pohl2012cognitive}.
\rev{Among various forms of reasoning,} \textit{syllogistic reasoning} is one of the basic forms of deductive reasoning and has been studied extensively~\cite{johnson1978psychology,evans1993human,Geurts2003-GEURWQ}.
However, the evaluation of LLMs and the construction of datasets incorporating these experimental methodologies has not been systematically pursued.

In this paper, \rev{we present the evaluation of LLMs with NeuBAROCO,} a manually constructed syllogism dataset with diverse properties and annotations designed to evaluate the reasoning abilities and biases of LLMs in syllogistic reasoning tasks.
NeuBAROCO is a bilingual (Japanese and English) dataset and includes detailed annotations for the types of reasoning biases associated with each problem.
The dataset is based on a problem set used for a series of psychological experiments assessing human reasoning ability with syllogisms~\cite{SHIKISHIMA2009256,shikishima2015genetic}.
A subset of the problems in the dataset is aligned with human performance metrics from these experiments.
Building on the work of \citet{ando-etal-2023-evaluating}, we have adapted this problem set to evaluate whether language models exhibit reasoning biases similar to those of humans.

\begin{table*}[t]
\centering
\scalebox{0.9}{
\begin{tabular}{lllll}  \toprule
\textbf{Type} & \textbf{Sentence Pattern} & \textbf{Predicate Logic} & \textbf{Set Theory} & \textbf{Description} \\ \midrule
\textbf{A} (\textit{all}) & \textsf{All $S$ are $P$} &
$\forall x (S(x) \to P(x))$ &
$S \subseteq P$ &
Universal Affirmative \\
\textbf{E} (\textit{no}) & \textsf{No $S$ are $P$} &
$\forall x (S(x) \to \neg P(x))$ &
$S \cap P = \emptyset$ &
Universal Negative \\
\textbf{I} (\textit{some}) & \textsf{Some $S$ are $P$} &
$\exists x (S(x) \wedge P(x))$ &
$S \cap P \neq \emptyset$ &
Particular Affirmative \\
\textbf{O} (\textit{some-not}) & \textsf{Some $S$ are not $P$} &
$\exists x (S(x) \wedge \neg  P(x))$ &
$S \setminus P \neq \emptyset$ &
Particular Negative  \\ \bottomrule
\end{tabular}
}%
\caption{Four types of categorical sentences and their translation into logical notations.}
\label{tab:categorical}
\end{table*}

The main contributions in this paper are summarized as follows.\footnote{
\rev{The data and results are available at \url{https://github.com/kmineshima/NeuBAROCO}.}
}
\rev{First, we constructed a dataset containing 95 and 790 syllogistic reasoning problems in the format of Multiple-Choice and Natural Language Inference (NLI) tasks, respectively.
This dataset design facilitates comparison with results and insights from human psychological experiments and preliminary studies on machine learning models.}

Second, we systematically investigated
various reasoning biases observed in LLMs in relation to the form and content of syllogisms, including \textit{belief biases}, \textit{conversion errors}, and \textit{atmosphere effects}, along with other tendencies, across multiple tasks and in both English and Japanese.

Finally, to more precisely identify the reasoning errors in LLMs, we introduce a new task called \textit{Translate-and-Explain}.
This task requires LLMs to first translate the syllogism into a logical expression (Translate) and then explain the reasoning process (Explain), by leveraging multi-step Chain-of-Thought prompting.
Our findings indicate two distinct error patterns: one in the interpretation of the syllogism demonstrated in translation, and the other in the explained reasoning process.
However, the primary limitations of these models lie in the reasoning process itself, rather than in the interpretation of syllogisms.

\rev{We first provide an overview of syllogistic reasoning in Section~\ref{sec:background}, followed by a detailed presentation of the dataset in Section~\ref{sec:dataset}. Subsequent sections present the evaluation tasks (Section~\ref{sec:tasks}) and the experimental setup along with an analysis of the results (Section~\ref{sec:experiment}). Additionally, Section~\ref{sec:related} reviews related work in the field.}

\section{Background: Syllogistic reasoning}
\label{sec:background}

In this study, we primarily focus on the logical inference task that classifies inferences into three labels: \textit{entailment}, \textit{contradiction}, and \textit{neutral} (neither entailment nor contradiction).

A syllogism is an inference that consists of two premises and one conclusion, where the premises and the conclusion are composed of four basic types of quantified sentences:
\textit{all}, \textit{no}, \textit{some}, and \textit{some-not}, as shown in Table~\ref{tab:categorical}.
For example, (1) and (2) are syllogisms composed of sentences with the quantifiers \textit{all} and \textit{no}.

\begin{center}
\begin{minipage}{9em}
\begin{center}
(1)
\syllogism{All B are C}
{All A are B}
{All A are C}
\end{center}
\end{minipage}
\hspace{1em}
\begin{minipage}{9em}
\begin{center}
(2)
\syllogism{No A are B}
{All B are C}
{No A are C}
\end{center}
\end{minipage}
\end{center}

\noindent
The syllogism in (1) is an instance of \textit{entailment}, where if the premises (\textbf{P1} and \textbf{P2}) are true, then the conclusion (\textbf{C}) is also true.
The syllogism in (2) is an instance of \textit{neutral}, where the relationship between the premises (\textbf{P1} and \textbf{P2}) and the conclusion (\textbf{C}) is neither entailment nor contradiction.

Syllogisms are relatively simple logical inferences that can be represented in monadic predicate logic~\cite{lukasiewicz1957},
a fragment of first-order logic where each predicate can take only one argument.
Table~\ref{tab:categorical} shows how to translate
each type of categorical sentence into logical notations,
that is, into predicate logic and set theory.

Despite being logically simple, syllogisms are known to be challenging inferences for humans.
Which types of syllogisms are prone to causing errors in human reasoning, or in other words, involve reasoning biases, is a topic widely studied in the field of cognitive science of human reasoning~\cite{evans1993human,manktelow1999reasoningEng,Geurts2003-GEURWQ,stenning2012human,khemlani2012theories}.
\rev{Our choice to focus on syllogistic reasoning is deliberate and aims to facilitate comparisons with insights from the extensive research on biases and reasoning in cognitive science.}
Some typical biases of syllogism will be introduced in Section~\ref{ssec:annotation}.
Focusing on the possibility of a detailed classification of such human reasoning biases, this study uses syllogistic reasoning in natural language as a testbed to evaluate the logical reasoning capabilities of LLMs.

\begin{table*}[t]
\small
\centering
\resizebox{\textwidth}{!}{
\begin{tabular}{lll} \toprule
\textbf{Type} & \textbf{English} & \textbf{Japanese} \\ \midrule
\multirow{3}{*}{\shortstack[l]{\textbf{Symbolic} \\ (98)}}
& \textbf{P1}: All A are B.
& \textbf{P1}: すべてのAはBである。\\
& \textbf{P2}: All B are C.
& \textbf{P2}: すべてのBはCである。\\
& \textbf{C}: All A are C.\hfill{(\textsc{Entailment})}
& \textbf{C}: すべてのAはCである。\hfill{(\textsc{Entailment})} \\
\midrule
\multirow{3}{*}{\shortstack[l]{\textbf{Congruent} \\(404)}}
& \textbf{P1}: All humans are mammals.
& \textbf{P1}: すべての人間は哺乳類である。\\
& \textbf{P2}: No mammal is a crocodile.
& \textbf{P2}: どの哺乳類もワニでない。\\
& \textbf{C}: No crocodile is human. \hfill{(\textsc{Entailment})}
& \textbf{C}: どのワニも人間でない。\hfill{(\textsc{Entailment})} \\ \midrule
\multirow{3}{*}{\shortstack[l]{\textbf{Incongruent} \\ (238)}}
& \textbf{P1}: Some animals are human beings.
& \textbf{P1}: ある動物は人間である。\\
& \textbf{P2}: All animals are tomatoes.
& \textbf{P2}: すべての動物はトマトである。\\
& \textbf{C}: Some humans are tomatoes. \hfill{(\textsc{Entailment})}
& \textbf{C}: ある人間はトマトである。\hfill{(\textsc{Entailment})} \\ \bottomrule
\end{tabular}
}%
\caption{
Examples of syllogisms in English and Japanese labeled as \textit{Symbolic}, \textit{Congruent}, and \textit{Incongruent}.
\rev{The numbers in parentheses indicate the number of cases for each type.}
}
\label{tab:neubaroco}
\end{table*}

\begin{table*}[t]
\small
\centering
\resizebox{\textwidth}{!}{
\begin{tabular}{l l l} \toprule
\textbf{Type} & \textbf{English} & \textbf{Japanese} \\ \midrule
\multirow{3}{*}{\shortstack[l]{\textbf{Conversion}\\ (66)}}
& \textbf{P1}: All B are A.
& \textbf{P1}: すべてのBはAである。\\
& \textbf{P2}: All B are C.
& \textbf{P2}: すべてのBはCである。\\
& \textbf{C}: All A are C.\hfill{(\textsc{Neutral})}
& \textbf{C}: すべてのAはCである。\hfill{(\textsc{Neutral})}\\
\midrule
\multirow{3}{*}{\shortstack[l]{\textbf{Atmosphere}\\ (345)}}
& \textbf{P1}: Some canines are pets.
& \textbf{P1}: 	あるイヌはペットである。\\
& \textbf{P2}: Some pets are smart.
& \textbf{P2}: あるペットは賢い。\\
& \textbf{C}: Some canines are smart.\hfill{(\textsc{Neutral})}
& \textbf{C}: あるイヌは賢い。\hfill{(\textsc{Neutral})}\\ \bottomrule
\end{tabular}
}%
\caption{
Examples of syllogisms labeled as \textit{Conversion} and \textit{Atmosphere}.
\rev{The numbers in parentheses indicate the number of cases for each type.}
}
\label{tab:neubaroco-bias}
\end{table*}

\section{The NeuBAROCO dataset}
\label{sec:dataset}

\rev{The NeuBAROCO dataset is based on a syllogism problem set called the BAROCO test~\cite{shikishima2005behavioural,SHIKISHIMA2009256}, originally designed for large-scale research on human reasoning abilities.}
BAROCO includes the so-called belief-bias tasks, which are typical examples involving human reasoning biases (see Section \ref{ssec:belief}).
In addition to linguistic tasks, it also includes Euler diagram tasks to test spatial cognition.
These formats of reasoning were used to investigate the correlation and the contributions of genetic and environmental factors through twin studies~\cite{shikishima2005behavioural}.
Furthermore, studies combining these tasks with experimental tasks in behavioral economics have also been conducted \cite{shikishima2015genetic}.

\citet{ando-etal-2023-evaluating} \rev{provides} a preliminary study preceding this research, aiming to apply the BAROCO problem set for evaluating LLMs by recasting a portion of the BAROCO problem set to introduce the NeuBAROCO dataset, consisting of 375 syllogistic reasoning problems.
In this study, we have expanded and utilized the NeuBAROCO dataset to evaluate the logical reasoning abilities of LLMs more systematically.
This expansion includes the addition of new problems, incorporating the Multiple-Choice problems \rev{(95 problems)} that were originally employed in the psychological experiments of the BAROCO \rev{problem} set.
\rev{Additionally, the number of NLI task problems has been expanded from 375 to 790.}

\subsection{Overview of the dataset}
\label{ssec:overview}

The original BAROCO problem set consists of two premises and multiple choices that could serve as conclusions.
\rev{Experiment participants} are asked to select a logically valid conclusion from the given choices.
The NeuBAROCO dataset was constructed by converting each problem from the BAROCO problem set into a format commonly used for the NLI task.
The NeuBAROCO dataset we use in this paper includes 790 problems for the NLI task, namely tasks that classify inferences into entailment, contradiction, and neutral.
Of these, 254 problems are classified as \textit{entailment}, 188 as \textit{contradiction}, and 348 as \textit{neutral}.
While the original BAROCO problem set is written entirely in Japanese, the NeuBAROCO dataset translates these problems into English, making it available as a Japanese-English parallel reasoning corpus.

\begin{figure}[]
\centering
\scalebox{0.9}{
\begin{tabular}{l}
\textbf{Premise 1}: All the rings in this box are Yuki's rings.\\
\textbf{Premise 2}: None of Yuki's rings are gold rings.\\
\ \\
1. All the rings inside this box are gold rings.\\
2. Some of the rings inside this box is a gold ring.\\
3. None of the rings in this box are gold rings.\\
4. Some ring inside this box is not a gold ring.\\
5. None of them.\\
\hfill{\textbf{Correct answer: 3}}
\end{tabular}
}%
\caption{An example of the Multiple-Choice task}
\label{tab:example-of-five-choice}
\end{figure}

The NeuBAROCO dataset also includes 80 problems for the Multiple-Choice task, along with 15 additional examples and practice problems, following the format of the original BAROCO problem set.
For each problem, the number of the correct answer is labeled.
Figure~\ref{tab:example-of-five-choice} shows an example of a problem in the Multiple-Choice task.

\subsection{Annotation}
\label{ssec:annotation}

\rev{We focus on three types of biases in syllogistic reasoning.
The three types of biases addressed here can be categorized into two kinds: biases related to \textit{content} (belief bias) and biases related to \textit{form} (conversion errors and atmosphere effects).
By investigating these three types of biases, we can systematically evaluate whether LLMs are sensitive to  the roles of content words and function words in deductive reasoning.}

\subsubsection{\rev{Labels for bias related to content}}
\label{ssec:belief}

To investigate biases caused by content words such as nouns and verbs,
we categorized each inference into three types based on whether it is congruent with commonsense beliefs.
Table~\ref{tab:neubaroco} shows examples and the number of instances for each type.

\paragraph{Symbolic}
When all terms are composed of sentences from abstract symbols,
the inference is labeled as \textit{symbolic}.
These types of problems are neutral with respect to the beliefs held by humans; that is, the question of whether they agree or disagree with those beliefs does not arise.

\paragraph{Congruent}

If there is no inconsistency with commonsense beliefs in all premises and conclusions, the inference is labeled as \textit{congruent}.

\paragraph{Incongruent}
If at least one of the premises or the conclusion does not align with commonsense beliefs, the inference is labeled as \textit{incongruent}.
In the example \rev{in the bottom row of} Table~\ref{tab:neubaroco}, \textit{All animals are tomatoes} and \textit{Some humans are tomatoes} contradict commonsense beliefs.

\medskip

If it is unclear whether the sentence is consistent with commonsense beliefs, or if it requires specialized knowledge (e.g., \textit{All agnostics are Stoics. Some agnostics are skeptical. Therefore, all Stoics are skeptical}), the inference is classified as \textit{others}. There are 50 instances of this type.

\subsubsection{\rev{Labels for bias related to form}}

In addition, we assigned the tags \textit{conversion} and \textit{atmosphere} to types of inferences that are prone to the two major types of reasoning biases.
\rev{These biases are induced by function words such as \textit{all} and \textit{not}, as well as by grammatical factors such as word order in the premises and conclusions.}

\paragraph{Conversion}

Conversion error is known as a typical reasoning bias in syllogisms \cite{evans1993human,Geurts2003-GEURWQ}.
This error occurs when quantified sentences are misinterpreted by converting the order of two terms:
\textit{All A are B} and \textit{Some A are not B} are misinterpreted as \textit{All B are A} and \textit{Some B are not A}, respectively.
For instance, interpreting the sentence \textit{All students who score above 90 points receive an A grade} as equivalent to \textit{All students with an A grade score above 90 points} exemplifies this error.
Although these two sentences may appear similar, they do differ in logical meaning.
Table~\ref{tab:neubaroco-bias} presents examples of syllogisms where such illicit conversion results in inference being erroneously classified as valid (\textit{entailment}) rather than invalid (\textit{neutral}).
\rev{We assign the \textit{conversion} label to those inferences where a sentence containing \textit{all} or \textit{some-not} appears in the premises, and the label changes from \textit{neutral} to \textit{entailment} when the order of terms in the sentence is reversed.}

\paragraph{Atmosphere}

\rev{The atmosphere effect indicates the tendency to select conclusions that mirror the form of the premises~\cite{woodworth1935atmosphere}.
This involves selecting conclusions that superficially resemble the premises in terms of their logical structure~\cite{chater1999probability}.
For example, a conclusion containing \textit{some} might be preferentially selected if a premise also contains \textit{some}. Similarly, if a premise containing a negation (\textit{no} or \textit{some-not}) tends to promote a negative conclusion.
We assign the \textit{atmosphere} label to those inferences
with \textit{neutral} labels where either (1) a premise contains \textit{some} and the conclusion is particular (\textit{some} or \textit{some-not}), or (2) a premise contains \textit{some-not} and the conclusion is particular or negative (i.e., \textit{no}, \textit{some}, or \textit{some-not}).
Table~\ref{tab:neubaroco-bias} shows an example that satisfies condition (1).}

\begin{table*}[t]
\centering
\scalebox{0.7}{
\begin{minipage}[t]{28em}
\begin{quoteboxa}{Input}
\tt

Select one statement from the five options provided that logically follows as a conclusion from the two premises presented in each problem. Answer by providing the number of your choice.

\medskip

Premise 1: All the rings in this box are Yuki's rings.\\
Premise 2: None of Yuki's rings are gold rings.

\medskip

1. All the rings inside this box are gold rings.\\
2. Some of the rings inside this box is a gold ring.\\
3. none of them.\\
4. None of the rings in this box are gold rings.\\
5. Some ring inside this box is not a gold ring.

\medskip

The answer is:
\end{quoteboxa}
\end{minipage}
}
\hspace{1em}
\scalebox{0.7}{
\begin{minipage}[t]{28em}
\begin{quoteboxa}{Input}
\tt

各問題にある2つの前提の結論として成り立つ文を、5つの選択肢の中から1つだけ選んでください。番号で回答してください。

\medskip

前提 1: この箱の中のすべての指輪はユキの指輪である。\\
前提 2: ユキのどの指輪も金の指輪でない。

\medskip

1. この箱の中のすべての指輪は金の指輪である。\\
2. この箱の中のある指輪は金の指輪である。\\
3. どれでもない。\\
4. この箱の中のどの指輪も金の指輪でない。\\
5. この箱の中のある指輪は金の指輪でない。

\medskip

答えは:
\end{quoteboxa}
\end{minipage}
}
\caption{Example prompts for the Multiple-Choice task in English and Japanese.}
\label{tab:prompt-choice5}
\end{table*}

\section{\rev{Evaluation Tasks}}
\label{sec:tasks}

We introduce three types of tasks for evaluating LLMs using the NeuBAROCO dataset: {Multiple-Choice}, {NLI}, and {Translate-and-Explain}.

\paragraph{Multiple-Choice}

\rev{The Multiple-Choice task is a format widely used in cognitive psychology. In this task, models are presented with two premises and asked to choose the correct conclusion from five options. These options include the \textit{all}, \textit{some}, \textit{no}, and \textit{some-not} sentences, as well as the ``none of them'' choice.
Table 4 presents examples of the prompts used in this task.}

\begin{table}[t]
\centering
\scalebox{0.68}{
\begin{minipage}[t]{29em}
\begin{quoteboxa}{Input}
\tt
Determine the correct logical relationship between the given premises and the hypothesis.\\
- Answer "entailment" if the hypothesis follows logically from the premises.\\
- Answer "contradiction" if the premises and the hypothesis are logically incompatible with each other.\\
- Answer "neither" if the relationship is neither "entailment" nor "contradiction".

\medskip

\#\# Input\\
Premise 1: Some A are B.\\
Premise 2: All B are C.\\
Hypothesis: All A are C.

\medskip

\#\# Translation into predicate logic\\ %
Premise 1: ∃x(Ax∧Bx)\\
Premise 2: ∀x(Bx→Cx)\\
Hypothesis: ∃x(Ax→Cx)

\medskip

\#\# Reasoning\\
{}[Explain your reasoning for the answer]

\medskip

\#\# Answer\\
{}[Your answer must be one word: "entailment", "contradiction", or "neither"]

\medskip

\#\# Input\\
Premise 1: One friend of Taro is a friend of Paul.\\
Premise 2: All of Paul's friends are German.\\
Hypothesis: All of Taro's friends are German.

\medskip

\#\# Translation into predicate logic
\end{quoteboxa}
\end{minipage}
}
\caption{Example prompt used for the Translate-and-Explain task (Translation into predicate logic).}
\label{tab:translate-example-in-section-input}
\end{table}

\paragraph{NLI}

The NLI task is a common problem setting in NLP, enabling evaluations on specific instances of reasoning and aligning well with other NLP benchmarks.
In this task, two premises of a syllogism and one hypothesis are presented. Models are then asked to determine whether the relationship of the hypothesis to the premises is one of \textit{entailment}, \textit{contradiction}, or \textit{neither}.
Experiments are conducted in zero-shot and few-shot settings, respectively~\cite{brown2020language}.
In the zero-shot setting, instructions and problems are provided without examples.
Example prompts are shown in Table~\ref{tab:prompt-choice3} of the Appendix.
In the few-shot (3-shot) setting, three exemplar problems with correct answers are included in the prompt (Table~\ref{tab:few-shot-examples} in Appendix).
\rev{We use abstract symbolic problems as exemplars to avoid potential biases, ensuring that the few-shot examples remain neutral to belief congruence and incongruence.}

\begin{table*}[t!]
\centering
\resizebox{0.7\textwidth}{!}{
\begin{tabular}{l l cS[table-format=2.2]cS[table-format=2.2]} \toprule
\textbf{Language} & \textbf{Model} & \textbf{Overall} & \textbf{Symbolic} & \textbf{Congruent} & \textbf{Incongruent}\\ \midrule
\multirow{6}{*}{English}
& GPT-3.5 & 53.75  & 	45.00 &	72.50 &	25.00\\
& GPT-4 & 83.75 &	75.00 &	90.00 &	80.00\\
& Llama-2-13B & 26.25 &	20.00 &	40.00 &	5.00\\
& Swallow-13B & 25.00 &	20.00 &	30.00 &	20.00\\
& Llama-2-70B & 56.25 & 65.00 &	62.50 &	35.00\\
& Swallow-70B & 60.00 & 45.00 &	80.00 &	35.00\\ \midrule
\multirow{7}{*}{Japanese}
& GPT-3.5 & 42.50  &	55.00 &	40.00 &	35.00\\
& GPT-4 & 95.00  &	85.00 &	97.50 &	100.00\\
& Llama-2-13B & 21.25 &	20.00 &	22.50 &	20.00\\
& Swallow-13B & 30.00 &	20.00 &	37.50 &	25.00\\
& Llama-2-70B & 66.25 &	60.00 &	62.50 &	80.00\\
& Swallow-70B & 50.00 & 45.00 &	52.50 &	50.00\\%
\cmidrule{2-6}
& \textbf{Human} & \rev{53.00} &	\rev{49.10} &	\rev{51.20} &	\rev{59.70}\\ \bottomrule

\end{tabular}
}
\caption{Accuracy (\%) on the Multiple-Choice task (80 problems).}
\label{tab:res-choice5}
\end{table*}

\paragraph{Translate-and-Explain}

To provide a finer-grained analysis of the reasoning ability of the models, we design the Translate-and-Explain task, a variant of the NLI task with a dedicated Chain-of-Thought (CoT) prompt.
\rev{In this task, we emulate the translation between natural language sentences and formal expressions of reasoning before the actual reasoning step, identifying whether errors and biases stem from the process of interpreting sentences or from the process of reasoning.}

CoT prompting is a technique of having LLMs perform intermediate reasoning steps by using few-shot examples or other means, and has been reported to improve the reasoning ability of LLMs \cite{wei2022chain}.
This enables us to apply a method analogous to the protocol analysis in psychological experiments \cite{evans1983conflict} to the evaluation of LLMs.

The Translate-and-Explain CoT prompt is a 1-shot structural prompt that instructs LLMs to perform (i) a translation step, (ii) an explanation step, and (iii) an answer step for each problem.
For the translation step, LLMs are instructed to translate the given syllogism into abstract expressions.
In this study, we compare translations into \textit{predicate logic} in formal language and into \textit{set-theory} in natural language (see Table~\ref{tab:categorical} in Section~\ref{sec:background}).
We conducted experiments in three setups: (a) explanation without translation, (b) predicate logic translation + explanation, (c) set-theoretic translation + explanation.
We manually checked the correctness of the translation outputs of LLMs.

Table~\ref{tab:translate-example-in-section-input} shows an example of the prompt in English.
Full examples are listed in Appendix~\ref{ssec:prompt-translate}.
\rev{
Note that we do not provide specific examples of reasoning in the prompt to avoid leading LLMs to adopt similar methods.
This allows us to analyze LLMs' ``free'' reasoning without conditioning.
}

\section{Experiments}
\label{sec:experiment}

\subsection{Experimental Setup}

We conducted experiments on the three types of tasks both in English and in Japanese.
In the Multiple-Choice task, we evaluated the following models:

\begin{itemize}[noitemsep,leftmargin=*]

\item
\textbf{GPT-3.5}~\cite{ouyang2022gpt3} and \textbf{GPT-4}~\cite{openai2023gpt4}.
The GPT models used were {gpt-3.5-turbo-1106} and {gpt-4-0613} available via OpenAI's API.
The number of parameters for the GPT models has not been disclosed.

\item
\textbf{Llama-2}~\cite{touvron2023llama} with 13 billion (13B) and 70 billion (70B) parameters.
For Llama-2, detailed model information, including the number of parameters, is publicly available.

\item
\textbf{Swallow}~\cite{fujii2024continual} with 13B and 70B parameters.
Swallow is a model family based on Llama-2 with state-of-the-art Japanese language capability, enhanced through continual pre-training using a dedicated corpus.
\end{itemize}

In the NLI and Translate-and-Explain tasks, the Llama-2 and Swallow models failed to produce output with a valid answer in most cases.
Therefore, we focus on the evaluation results of the GPT models, which consistently produced valid outputs.

The default values were used for the hyperparameters, except for the maximum output token length.
This was set to 10 for the Multiple-Choice and NLI tasks, and 2,048 for the Translate-and-Explain task, which are sufficiently long given the design of the tasks.

\begin{table*}[t]
\centering
\resizebox{0.9\textwidth}{!}{
\begin{tabular}{l l S[table-format=2.2] S[table-format=2.2]S[table-format=2.2]S[table-format=2.2] S[table-format=2.2]S[table-format=2.2]S[table-format=2.2]S[table-format=2.2]S[table-format=2.2]} \toprule
\textbf{Language} & \textbf{Model} & \textbf{Overall} & {\textbf{E}} & {\textbf{C}} & {\textbf{N}} & \textbf{Symbolic} & \textbf{\rev{Cong}} & \textbf{\rev{Incong}} & \textbf{Conv} & \textbf{Atmos} \\ \midrule
\multirow{4}{*}{English}
& GPT-3.5 & 49.75 &	84.25 &	42.02 &	28.74 & 55.10 & 56.19 & 36.55 & 18.18 & 32.67\\
& \quad (Few-Shot) & 47.09 &	88.58 &	27.66 &	27.30 & 47.96 & 50.50 & 39.92 & 13.64 & 29.70\\ %
& GPT-4 & 71.77 &	85.04 &	93.62 &	50.29 & 76.53 & 76.24 & 61.76 & 40.91 & 50.99\\
& \quad (Few-Shot) & 77.47 &	90.55 &	88.83 &	61.78 & 79.59 & 82.92 & 67.23 & 50.00 & 59.41\\ \midrule
\multirow{4}{*}{Japanese}
& GPT-3.5 & 40.00 &	82.28 &	51.06 &	3.16 & 35.71 & 48.02 & 28.57 & 4.55 & 1.98\\
& \quad (Few-Shot) & 40.00 &	90.94 &	35.64 &	5.17 & 35.71 & 46.53 & 31.09 & 4.55 & 5.45\\ %
& GPT-4 & 70.38 &	87.40 &	95.74 &	44.25 & 71.43 & 76.73 & 60.92 & 43.94 & 38.12\\
& \quad (Few-Shot) & 78.61 &	92.13 &	88.30 &	63.51 & 82.65 & 81.44 & 73.95 & 74.24 & 57.43\\ \bottomrule
\end{tabular}
}%
\caption{Accuracy (\%) on the NLI task (790 problems).
\textbf{E} = \textit{entailment},
\textbf{C} = \textit{contradiction},
\textbf{N} = \textit{neutral},
\textbf{Cong} = \textit{Congruent},
\textbf{Incong} = \textit{Incongruent},
\textbf{Conv} = \textit{Conversion},
\textbf{Atmos} = \textit{Atmosphere}.
}
\label{tab:res-choice3}
\end{table*}

\subsection{Results and Analysis}
\label{ssec:results}

\subsubsection{Multiple-Choice Task}
\label{sssec:mc-results}

Table~\ref{tab:res-choice5} shows the experimental results for the Multiple-Choice task.
The row labeled \textbf{Human} presents the average scores of 440 participants based on data from the psychological experiment conducted by \citet{SHIKISHIMA2009256}.
Note that there are some terminological differences between \citet{SHIKISHIMA2009256} and our study.

\rev{In terms of the biases related to content, \textit{incongruent} cases are generally harder for the LLMs than \textit{congruent} cases in English.
In contrast, no similar trend is clearly observed in Japanese problems.}

In terms of model scale, while the overall accuracy of the smaller (13B-parameter) models ranged from 20\% (\rev{chance level}) to 30\%, the 70B-parameter models and GPTs achieved an overall accuracy of 42\% to 95\%, with some surpassing human overall accuracy in Japanese (53\%).
\rev{
The strong performance of large-parameter LLMs can be partly attributed to the nature of the Multiple-Choice task.
As we will see in Section~\ref{sssec:NLI-results}, LLMs (especially those with larger parameter) mark high accuracy particularly in identifying \textit{entailment} and \textit{contradiction} over \textit{neutral} cases.
In this task, the correct choice is always an \textit{entailment} of the given premises, unless the answer is ``none of them.''
Following the original BAROCO problem set, the NeuBAROCO dataset does not include any problems where the correct answer is ``none of them.''
}

\rev{
While Swallow is a Japanese-enhanced model based on Llama-2, the comparative results for the two are not straightforward.
For a detailed discussion, see Appendix~\ref{sec:detailed-results}.
}

\subsubsection{NLI Task}
\label{sssec:NLI-results}

\begin{table*}[t]

\centering
\scalebox{0.85}{
\begin{tabular}{l l c cc cc}
\toprule
& & \textbf{Explanation Only} & \multicolumn{2}{c}{\textbf{Pred + Explanation}} & \multicolumn{2}{c}{\textbf{Set + Explanation}} \\ \cmidrule(lr){3-3} \cmidrule(lr){4-5} \cmidrule(lr){6-7}
\textbf{Language} & \textbf{Model} & \textbf{Answer} &
\textbf{Translation} & \textbf{Answer} & \textbf{Translation} & \textbf{Answer} \\ \midrule
\multirow{2}{*}{English}
& GPT-3.5 & 58.89 & 91.11 & 52.22 & 87.78 & 54.44 \\
& GPT-4 & 77.78 & 96.67 & 83.33 & 92.22 & 80.00 \\ \midrule
\multirow{2}{*}{Japanese}
& GPT-3.5 & 57.78 & 71.11 & 57.78 & 33.33 & 60.00 \\
& GPT-4 & 70.00 & 95.56 & 75.56 & 76.66 & 74.44 \\ \bottomrule
\end{tabular}
} %
\caption{Accuracy (\%) on the Translate-and-Explain task (90 problems).
\textbf{Pred} = \textit{Predicate logic},
\textbf{Set} = \textit{Set theory}.
}
\label{tab:translation-task}
\end{table*}

Table~\ref{tab:res-choice3} shows the results of the NLI task.
The few-shot setting improves the overall accuracy of GPT-4 both in English and Japanese.

In terms of gold labels, The models achieve higher scores on problems labeled as \textit{entailment}, while those labeled as \textit{neutral} are typically the most challenging.
Even GPT-4, which performed the best, scored approximately 30 points lower on \textit{neutral} problems compared to other labels.
\rev{
As noted above, when comparing the Multiple-Choice and NLI tasks, the Multiple-Choice task includes only problems that correspond to the \textit{entailment} problems in the NLI task.
Consequently, the scores for the Multiple-Choice task are similar to those for the \textit{entailment} problems in the NLI task.
}

In terms of bias-related labels, the results suggest that LLMs are influenced by the content and form of syllogisms.
Among the \textit{symbolic}, \textit{congruent}, and \textit{incongruent} cases, the accuracies in the incongruent cases are generally lower than those in the other cases and the overall score for each model.
Regarding conversion errors, the accuracies for the problems labeled \textit{conversion}, which may cause conversion errors, are significantly lower than the overall accuracies.
A similar trend is observed for the problems labeled \textit{atmosphere} in the context of atmosphere effects.

\subsubsection{Translate-and-Explain Task}
\label{sssec:TE-results}

Table~\ref{tab:translation-task} shows the experimental results for the Translate-and-Explain task.
Table~\ref{tab:translate-example-in-section-output} provides an example output from GPT-4.

\begin{table}[t]
\centering
\scalebox{0.7}{
\begin{minipage}[t]{28em}
\begin{quoteboxa}{Output (GPT-4)}
\tt
Premise 1: ∃x((Friend(Taro, x) ∧ Friend(Paul, x))\\ %
Premise 2: ∀x((Friend(Paul, x) → ¬German(x))\\
Hypothesis: ∃x(Friend(Taro, x) ∧ German(x))

\medskip

\#\# Reasoning\\
Given that one friend of Taro is also a friend of Paul's and none of Paul's friends are German, it follows that this mutual friend cannot be German. Thus, the hypothesis that one of Taro's friends is German is not contradictory to any of the premises, because while we know one friend is not German, it is still possible for Taro to have other friends who are German.

\medskip

\#\# Answer\\
neither

\end{quoteboxa}
\end{minipage}
}
\caption{Example \rev{response} from GPT-4 in the Translate-and-Explain task (Translation into predicate logic).}
\label{tab:translate-example-in-section-output}
\end{table}

\paragraph{English}

\rev{With} GPT-4, the translations from English to predicate logic are highly accurate, correctly translating 87/90 problems.
However, translating the same problems into set theory is more challenging, with 82/90 problems correctly translated.
It is observed that GPT-4 often responds with \textit{contradiction} when the correct answer to an inference is \textit{neither}.
As a typical error, GPT-4 interpreted \textit{All animals are tomatoes} as \textit{The set of animals is identical to the set of tomatoes}
rather than \textit{The set of animals is a subset of the set of tomatoes}.

\rev{With} GPT-3.5, the translations into predicate logic are almost accurate, correctly translating 82/90 problems.
As a typical error, GPT-3.5 interprets
\textit{A certain police officer is not a public servant} as $\neg \exists x(Px \wedge Sx)$ instead of $\exists x(Px \wedge \neg
 Sx)$,
failing to capture the correct scope of negation.
The number of problems translated correctly to set theory is 79/90.
GPT-3.5 mistakenly interprets
\textit{A certain police officer is not a public servant}
as \textit{The set of police officers does not intersect with the set of public servants},
which is logically equivalent to the wrong predicate logic formula, $\neg \exists x(Px \wedge Sx)$.

\paragraph{Japanese}

\rev{With} GPT-4, 86/90 problems are correctly translated from Japanese to predicate logic.
Translations into set theory are more error-prone than those into predicate logic, with only 69/90 problems correctly translated.
GPT-3.5 has trouble in translation into predicate logic, with 64/90 problems correctly translated.
Similar to the mistakes often made by beginners in logic, the model incorrectly interprets a Japanese sentence corresponding to
\textit{Some A are B} as $\exists x (A(x) \to B(x))$, while the correct interpretation is $\exists x (A(x) \wedge B(x))$.

\rev{With} GPT-3.5, 28/90 problems are correctly translated from Japanese to set theory.
The translation of A-type (\textit{all}-type) sentences in Japanese
is generally accurate, but the translations of the sentences of other types
tend to include mistakes. Especially, O-type (\textit{some-not})
and E-type (\textit{no}) sentences are frequently confused.
For example, a Japanese sentence corresponding to
\textit{Some robots are not commodities} is incorrectly interpreted as
\textit{The set of products and the set of robots do not have any common elements}.

\section{Related Work}
\label{sec:related}

Recent years have seen active research on machine learning and deep learning models that focus on logical reasoning in natural language, including syllogistic reasoning.

\citet{richardson2020probing} investigated the abilities of NLI models in handling various logical inferences involving boolean operators, quantifiers, conditionals, and negation using synthetically generated data.
\citet{yanaka2019can} studied monotonicity inferences.
Monotonicity inferences are simpler than syllogistic inferences in that they only have single premises, whereas syllogisms involve multiple premises with challenging combinations of quantifiers and negation.
\citet{schlegel-etal-2022-transformers} conducted an empirical study to explore the detection of formally valid inferences within controlled fragments of natural language, designed to increase satisfiability problem complexity.
These studies combine pre-training and fine-tuning with relatively large datasets for logical reasoning and are not aimed at evaluating current LLMs based on in-context learning.

As datasets for learning and evaluating syllogistic reasoning, \citet{dong2020learning} and \citet{gubelmann2022philosophically} have constructed datasets using linguistic resources such as WordNet and their own lists of words, employing template-based automatic generation methods based on types of syllogisms.
The dataset of \citet{gubelmann2022philosophically} includes
three-class labels (entailment, contradiction, and neutral) but does not include additional information related to reasoning biases.
AVICENNA~\cite{aghahadi2022avicenna} is a crowdsourced dataset containing binary labels that indicate whether the conclusion of a syllogism follows from two given sentences (and if so, the conclusion sentence is also provided), but it does not contain information about reasoning biases.
SYLLOBASE~\cite{wu2023hence} contains five types of syllogisms.
It consists 50,000 syllogism samples automatically generated from existing knowledge bases, with 1,000 of these samples manually annotated as a test set.
Experiments were conducted in zero-shot and few-shot settings, covering both generation and selection tasks.
All of the datasets above are in English.

\rev{As a framework for generating deduction datasets, FLD \cite{pmlr-v202-morishita23a} and its Japanese version, JFLD \cite{morishita-etal-2024-jfld-japanese}, have been proposed.
\citet{pmlr-v202-morishita23a} empirically verifies that language models trained with FLD demonstrate enhanced generalizable deductive reasoning abilities.}

\citet{dasgupta2022language} and \citet{eisape2023systematic} are the works closely related to ours.
\citet{dasgupta2022language} \rev{focused} on belief biases in \rev{syllogistic reasoning} and showed that the alignment of conclusions with human beliefs affects the performance of \rev{LLMs}.
In their studies, scenarios were classified based on whether the content of the conclusion contradicts our beliefs, does not contradict, or consists of meaningless words.
LLMs were then tasked with determining whether a combination of two premises and one conclusion constitutes valid or invalid syllogistic reasoning in a binary choice format.
This approach differs from the three-class classification used in our work.
Our research aims to further explore reasoning biases in LLMs by examining various types of biases, particularly those related not only to the content of reasoning, as highlighted by belief bias, but also to the form of reasoning.
Additionally, we investigated these biases in experimental settings, including the Multiple-Choice and Translate-and-Explain tasks.
\rev{
\citet{eisape2023systematic}, focusing on syllogisms, investigates how language models perform logical reasoning compared to humans.
Experiments were conducted using the PaLM 2 models, and the results indicate that larger models perform more accurately than smaller models and humans.
It was also confirmed that even the largest models tend to make errors that reflect human reasoning biases in certain types of syllogisms.
}

\rev{\citet{ando-etal-2023-evaluating} is a preliminary work that precedes our research, reporting only the zero-shot performance of GPT-3.5 across 375 NLI tasks. In contrast, our study reports results on 790 NLI tasks, 80 Multiple-Choice tasks using Chain-of-Thought prompts, and 90 Translate-and-Explain tasks. Furthermore, we evaluated a broader range of models, including GPT-4, Llama-2 (13B and 70B), and Swallow (13B and 70B), in both zero-shot and few-shot settings. The findings related to the Multiple-Choice and Translate-and-Explain tasks represent entirely new contributions of this study.}

\section{Summary and Conclusion}
\label{sec:conclusion}

In this paper, \rev{we evaluated} the logical reasoning \rev{ability} of LLMs by developing the NeuBAROCO dataset, which consists of syllogisms in both Japanese and English,
annotated with information on reasoning biases and results from large-scale human evaluation data.
The results of experiments using the NLI task demonstrated that some LLMs, particularly GPT-4, achieved high accuracy in both English and Japanese, especially in \rev{the few-shot setting.}
However, significant room for improvement remains in problems where the entailment relation is neutral (neither entailment nor contradiction). The results suggest that, particularly in problems labeled as \textit{Conversion} and \textit{Atmosphere}, there is a tendency to exhibit the same reasoning biases as humans.

Moreover, in the experiments using the Multiple-Choice task, a method commonly used in psychological studies of syllogisms, some models achieved accuracies surpassing those of human participants in large-scale experiments. It is important to note that the methodology for comparing human accuracy with LLMs is not yet fully established, and a detailed comparison with human performance remains a future challenge.

Finally, in the Translate-and-Explain task, which requires providing explanations of reasoning along with translations of syllogisms into logical expressions, many models showed improved accuracy, with translation accuracy nearly reaching 100\%. However, reasoning errors remain, suggesting that the source of these errors is not the misinterpretation of the premise sentences but rather the reasoning process itself.

While syllogisms represent one of the basic forms of logical reasoning used in psychology, expanding our evaluation to include more diverse and complex natural language inferences that induce reasoning biases remains an essential challenge.
This includes boolean propositional inferences~\cite{evans1989bias},
\textit{if-then} conditionals~\cite{johnson2002conditionals},
and spatial inferences involving polyadic relations~\cite{byrne1989spatial}.
\rev{The biases we addressed are not specific to syllogistic reasoning and, therefore, may potentially be generalized to tasks beyond syllogistic reasoning.}
Addressing these issues in future research is crucial for advancing our understanding of the reasoning capabilities of LLMs.

\section*{Limitations}

In this study, we used closed-source LLMs, particularly noting that the GPT models we used lack precise information about the models, such as the number of parameters, and scale and distribution of the training data.
Although the NeuBAROCO dataset has not been publicly released so far, it cannot be denied that some parts of syllogistic reasoning, especially symbolic syllogisms themselves, may be included in the training data.
Also, there is a risk associated with using closed-source models in scientific research due to the lack of reproducibility.

Comparing the accuracy of LLMs to human accuracy obtained from psychological experiments presents an intriguing challenge for research in both NLP and cognitive psychology regarding human reasoning with natural languages.
However, to our knowledge, there is not yet a well-established methodology for comparing the accuracy of LLMs with that of humans yet.
In the case of the BAROCO project, which involves more than 400 participants, further discussion is necessary on how to systematically and meaningfully compare the accuracy of LLMs with that of humans.

Research on English and Japanese \rev{LLMs} is rapidly advancing even as our study progresses.
While evaluations are being conducted using a variety of current representative models, both open and closed, it is naturally impossible to cover all of them comprehensively.
Particularly, the number of Japanese LLMs is increasing, and conducting a systematic comparison of Japanese and English LLMs is one of the important challenges for the future.

\section{Acknowledgements}

\rev{We thank the anonymous reviewers for their insightful comments and suggestions, which have greatly improved the paper. We also extend our gratitude to Professors Juko Ando and Chizuru Shikishima, as well as the members of their KTS-KoTReC project, for their collaboration with the last author in developing the original BAROCO test set and for providing us with the associated human subject data.
This work is partially supported by JST, CREST Grant Number JPMJCR2114 and JSPS Kakenhi Grant Numbers JP24K00004, JP21K00016, JP21H00467, JP23K20416, and JP21K18339.}

\bibliography{custom}
\bibliographystyle{acl_natbib}

\appendix

\section{Examples of Prompts}

\subsection{Multiple-Choice task}

Table~\ref{tab:prompt-choice5} in Section \ref{sec:experiment} shows examples of prompts for the Multiple-Choice task.

\subsection{NLI task}

Table~\ref{tab:prompt-choice3} shows examples of the prompts we used for the NLI task in English and Japanese.
We employed the most effective prompt pattern from among several we evaluated, including those listed in Table~\ref{tab:prompt-alternatives}.
Table~\ref{tab:few-shot-examples} shows examples of few-shot prompts for the task.

\begin{table*}[hp]
\centering
\scalebox{0.7}{
\begin{minipage}[t]{28em}
\begin{quoteboxa}{Input}
\tt
Determine the correct logical relationship between the given premises and the hypothesis.\\
- Answer ``entailment'' if the hypothesis follows logically from the premises.\\
- Answer ``contradiction'' if the premises and the hypothesis are logically incompatible with each other.\\
- Answer ``neither'' if the relationship is neither ``entailment'' nor ``contradiction''.\\
Your answer must be one word: ``entailment'', ``contradiction'', or ``neither''.\\[\normalbaselineskip]
Premise 1: One friend of Taro is a friend of Paul.\\
Premise 2: All of Paul's friends are German.\\
Hypothesis: All of Taro's friends are German.\\
The answer is:
\end{quoteboxa}
\end{minipage}
}
\hspace{1em}
\scalebox{0.7}{
\begin{minipage}[t]{28em}
\begin{quoteboxa}{Input}
\tt
与えられた前提と仮説の正しい論理的関係を判定しなさい。\\
- 仮説が前提から論理的に導かれる場合は「含意」と答えなさい。\\
- 前提と仮説が論理的に両立しない場合は「矛盾」と答えなさい。\\
- その関係が「含意」でも「矛盾」でもない場合は「どちらでもない」と答えなさい。\\
「含意」「矛盾」「どちらでもない」のいずれか一語で回答しなさい。\\[\normalbaselineskip]
前提1: 太郎のある友人はポールの友人である。\\
前提2: ポールのすべての友人はドイツ人である。\\
仮説: 太郎のすべての友人はドイツ人である。\\
答えは:
\end{quoteboxa}
\end{minipage}
}
\caption{Example prompts for the NLI task.}
\label{tab:prompt-choice3}
\end{table*}

\begin{table*}[hp]
\centering
\scalebox{0.7}{
\begin{minipage}[t]{28em}
\begin{quoteboxa}{Input}
\tt
Carefully evaluate the following inference, and determine whether the premises entail or contradict the conclusion.
Answer exactly with entailment, contradiction, or neither.
The answer may be entailment even if it contradicts common sense.
For example, the answer to the following inference is entailment.
\\
[\normalbaselineskip]
Premise 1: All people are watermelons.\\
Premise 2: All watermelons are delicious.\\
Conclusion: All people are delicious.\\
\end{quoteboxa}
\end{minipage}
}
\hspace{1em}
\scalebox{0.7}{
\begin{minipage}[t]{28em}
\begin{quoteboxa}{Input}
\tt
Carefully evaluate the following inference, and determine whether the premises entail or contradict the conclusion.
Answer exactly with entailment, contradiction, or neither.
Let's think step by step.

\end{quoteboxa}
\end{minipage}
}
\caption{Examples of other tested prompts not adopted.}
\label{tab:prompt-alternatives}
\end{table*}

\begin{table*}[hp]
\centering
\scalebox{0.7}{
\begin{minipage}[t]{28em}
\begin{quoteboxa}{Input}
\tt
Determine the correct logical relationship between the given premises and the hypothesis.\\
- Answer ``entailment'' if the hypothesis follows logically from the premises.\\
- Answer ``contradiction'' if the premises and the hypothesis are logically incompatible with each other.\\
- Answer ``neither'' if the relationship is neither ``entailment'' nor ``contradiction''.\\
Your answer must be one word: ``entailment'', ``contradiction'', or ``neither''.

\medskip

Premise 1: Some X are Y.\\
Premise 2: All Y are Z.\\
Hypothesis: All X are Z.\\
The answer is: neither

\medskip

Premise 1: Some X are Y.\\
Premise 2: All Y are Z.\\
Hypothesis: Some X are Z.\\
The answer is: entailment

\medskip

Premise 1: Some X are Y.\\
Premise 2: All Y are Z.\\
Hypothesis: No X are Z.\\
The answer is: contradiction

\medskip

Premise 1: One friend of Taro is a friend of Paul.\\
Premise 2: All of Paul's friends are German.\\
Hypothesis: All of Taro's friends are German.\\
The answer is:
\end{quoteboxa}
\end{minipage}
}
\hspace{1em}
\scalebox{0.7}{
\begin{minipage}[t]{28em}
\begin{quoteboxa}{Input}
\tt
与えられた前提と仮説の正しい論理的関係を判定しなさい。\\
- 仮説が前提から論理的に導かれる場合は「含意」と答えなさい。\\
- 前提と仮説が論理的に両立しない場合は「矛盾」と答えなさい。\\
- その関係が「含意」でも「矛盾」でもない場合は「どちらでもない」と答えなさい。\\
「含意」「矛盾」「どちらでもない」のいずれか一語で回答しなさい。

\medskip

前提1: あるXはYである。\\
前提2: すべてのYはZである。\\
仮説: すべてのXはZである。\\
答えは: どちらでもない

\medskip

前提1: あるXはYである。\\
前提2: すべてのYはZである。\\
仮説: あるXはZである。\\
答えは: 含意

\medskip

前提1: あるXはYである。\\
前提2: すべてのYはZである。\\
仮説: どのXもZでない。\\
答えは: 矛盾

\medskip

前提1: 太郎のある友人はポールの友人である。\\
前提2: ポールのすべての友人はドイツ人である。\\
仮説: 太郎のすべての友人はドイツ人である。\\
答えは:

\end{quoteboxa}
\end{minipage}
}
\caption{Example prompts for the NLI task in the few-shot setting.}
\label{tab:few-shot-examples}
\end{table*}

\subsection{Translate-and-Explain task}
\label{ssec:prompt-translate}

Table~\ref{tab:prompt-translate-fol} and Table~\ref{tab:prompt-translate-set} show examples of the prompts we used for the Translate-and-Explain task in English and Japanese.

\begin{table*}[hp]
\centering
\scalebox{0.7}{
\begin{minipage}[t]{28em}
\begin{quoteboxa}{Input}
\tt
Determine the correct logical relationship between the given premises and the hypothesis.\\
- Answer "entailment" if the hypothesis follows logically from the premises.\\
- Answer "contradiction" if the premises and the hypothesis are logically incompatible with each other.\\
- Answer "neither" if the relationship is neither "entailment" nor "contradiction".

\medskip

\#\# Input\\
Premise 1: Some A are B.\\
Premise 2: All B are C.\\
Hypothesis: All A are C.

\medskip

\#\# Translation to predicate logic\\
Premise 1: ∃x(Ax∧Bx)\\
Premise 2: ∀x(Bx→Cx)\\
Hypothesis: ∀x(Ax→Cx)

\medskip

\#\# Reasoning\\
{}[Explain your reasoning for the answer]

\medskip

\#\# Answer\\
{}[Your answer must be one word: "entailment", "contradiction", or "neither"]

\medskip

\#\# Input\\
Premise 1: One friend of Taro is a friend of Paul.\\
Premise 2: All of Paul's friends are German.\\
Hypothesis: All of Taro's friends are German.

\medskip

\#\# Translation to predicate logic

\end{quoteboxa}

\begin{quoteboxa}{Output (GPT-4)}
\tt
Premise 1: ∃x((Friend(Taro, x) ∧ Friend(Paul, x))\\
Premise 2: ∀x((Friend(Paul, x) → ¬German(x))\\
Hypothesis: ∃x(Friend(Taro, x) ∧ German(x))

\medskip

\#\# Reasoning\\
Given that one friend of Taro is also a friend of Paul's and none of Paul's friends are German, it follows that this mutual friend cannot be German. Thus, the hypothesis that one of Taro's friends is German is not contradictory to any of the premises, because while we know one friend is not German, it is still possible for Taro to have other friends who are German.

\medskip

\#\# Answer\\
neither

\end{quoteboxa}

\end{minipage}
}
\hspace{1em}
\scalebox{0.7}{
\begin{minipage}[t]{28em}
\begin{quoteboxa}{Input}
\tt
与えられた前提と仮説の正しい論理的関係を判定しなさい。\\
- 仮説が前提から論理的に導かれる場合は「含意」と答えなさい。\\
- 前提と仮説が論理的に両立しない場合は「矛盾」と答えなさい。\\
- その関係が「含意」でも「矛盾」でもない場合は「どちらでもない」と答えなさい。

\medskip

\#\# 入力\\
前提1: あるAはBである。\\
前提2: すべてのBはCである。\\
仮説: すべてのAはCである。\\

\#\# 述語論理への翻訳\\
前提1: ∃x(Ax∧Bx)\\
前提2: ∀x(Bx→Cx)\\
仮説: ∀x(Ax→Cx)

\medskip

\#\# 論証\\
{}[答えを導く論証を説明しなさい]

\medskip

\#\# 答え\\
{}[「含意」「矛盾」「どちらでもない」のいずれか一語で回答しなさい]

\medskip

\#\# 入力\\
前提1: 太郎のある友人はポールの友人である。\\
前提2: ポールのどの友人もドイツ人でない。\\
仮説: 太郎のある友人はドイツ人である。

\medskip

\#\# 述語論理への翻訳
\end{quoteboxa}

\begin{quoteboxa}{Output (GPT-4)}
\tt
前提1: ∃x(Fxt∧Fpx)\\
前提2: ∀x(Fpx→\textasciitilde Gx)\\
仮説: ∃x(Fxt∧Gx)

\medskip

\#\# 論証\\
前提1と前提2から、「x」を「ポールの友人であり太郎の友人でもある人」に置き換えると、その人はドイツ人ではない（\textasciitilde Gx）と結論付けることができる。これは仮説、すなわち「太郎のある友人はドイツ人である」と矛盾している。

\medskip

\#\# 答え\\
矛盾

\end{quoteboxa}

\end{minipage}
}
\caption{Example prompts used for the Translate-and-Explain task (translation to predicate logic).}
\label{tab:prompt-translate-fol}
\end{table*}

\begin{table*}[hp]
\centering
\scalebox{0.7}{
\begin{minipage}[t]{28em}
\begin{quoteboxa}{Input}
\tt
Determine the correct logical relationship between the given premises and the hypothesis.\\
- Answer "entailment" if the hypothesis follows logically from the premises.\\
- Answer "contradiction" if the premises and the hypothesis are logically incompatible with each other.\\
- Answer "neither" if the relationship is neither "entailment" nor "contradiction".

\medskip

\#\# Input\\
Premise 1: Some A are B.\\
Premise 2: All B are C.\\
Hypothesis: All A are C.

\medskip

\#\# Translation to set theory\\
Premise 1: The set of As overlaps with the set of Bs.\\
Premise 2: The set of Bs is a subset of the set of Cs.\\
Hypothesis: The set of As is a subset of the set of Cs.

\medskip

\#\# Reasoning\\
{}[Explain your reasoning for the answer]

\medskip

\#\# Answer\\
{}[Your answer must be one word: "entailment", "contradiction", or "neither"]

\medskip

\#\# Input\\
Premise 1: One friend of Taro is a friend of Paul's.\\
Premise 2: None of Paul's friends are German.\\
Hypothesis: One of Taro's friends is German.

\medskip

\#\# Translation to set theory
\end{quoteboxa}

\begin{quoteboxa}{Output (GPT-4)}
\tt
Premise 1: The set of Taro's friends overlaps with the set of Paul's friends.\\
Premise 2: The set of Paul's friends is disjoint from the set of Germans.\\
Hypothesis: The set of Taro's friends overlaps with the set of Germans.

\medskip

\#\# Reasoning\\
The first premise tells us that Taro and Paul share at least one friend. The second premise tells us that none of Paul's friends are German. Therefore, since the friend they share can't be German, the hypothesis that one of Taro's friends is German doesn't necessarily follow from the premises nor contradicts them, since Taro could have other friends that aren't friends with Paul.

\medskip

\#\# Answer\\
Neither
\end{quoteboxa}

\end{minipage}
}
\hspace{1em}
\scalebox{0.7}{
\begin{minipage}[t]{28em}
\begin{quoteboxa}{Input}
\tt
与えられた前提と仮説の正しい論理的関係を判定しなさい。\\
- 仮説が前提から論理的に導かれる場合は「含意」と答えなさい。\\
- 前提と仮説が論理的に両立しない場合は「矛盾」と答えなさい。\\
- その関係が「含意」でも「矛盾」でもない場合は「どちらでもない」と答えなさい。

\medskip

\#\# 入力\\
前提1: あるAはBである。\\
前提2: すべてのBはCである。\\
仮説: すべてのAはCである。

\medskip

\#\# 集合論への翻訳\\
前提1: Aの集合とBの集合は共通部分を持つ。\\
前提2: Bの集合はCの集合の部分集合である。\\
仮説: Aの集合はCの集合の部分集合である。

\medskip

\#\# 論証\\
{}[答えを導く論証を説明しなさい]

\medskip

\#\# 答え\\
{}[「含意」「矛盾」「どちらでもない」のいずれか一語で回答しなさい]

\medskip

\#\# 入力\\
前提1: 太郎のある友人はポールの友人である。\\
前提2: ポールのどの友人もドイツ人でない。\\
仮説: 太郎のある友人はドイツ人である。

\medskip

\#\# 集合論への翻訳
\end{quoteboxa}

\begin{quoteboxa}{Output (GPT-4)}
\tt
前提1: 太郎の友人の集合とポールの友人の集合は共通部分を持つ。\\
前提2: ポールの友人の集合はドイツ人の集合と共通部分を持たない。\\
仮説: 太郎の友人の集合とドイツ人の集合は共通部分を持つ。

\medskip

\#\# 論証\\
太郎のある友人がポールの友人であり、ポールのすべての友人がドイツ人でないという前提から、太郎のすべての友人がドイツ人でないという結論を導くことができる。したがって、仮説は前提と論理的に矛盾している。

\medskip

\#\# 答え\\
矛盾

\end{quoteboxa}

\end{minipage}
}
\caption{Example prompts used for the Translate-and-Explain task (translation to set theory).}
\label{tab:prompt-translate-set}
\end{table*}

\section{Llama-2 and Swallow comparison}
\label{sec:detailed-results}

\paragraph{Llama-2 and Swallow comparison.}
In the Multiple-Choice task, Llama-2-13B performs better in English than in Japanese, while Llama-2-70B performs better in Japanese than in English.
Conversely, Swallow-13B excels in Japanese, while Swallow-70B excels in English.
These results indicate that performance trends can change significantly within the same model family depending on the model scale, at least for inference tasks.
Furthermore, when comparing the two, Swallow-13B outperforms Llama-2-13B in Japanese but underperforms in English, whereas Swallow-70B underperforms Llama-2-70B in Japanese but outperforms it in English.
Although Swallow is a model family that improves Llama-2 through continual pre-training on a large-scale Japanese corpus, continual pre-training in a specific language does not always enhance performance in that language for inference tasks, indicating a complex effect.

\end{document}